\documentclass[lettersize,journal]{IEEEtran}

\usepackage{amsmath,amsfonts}
\usepackage{mathtools}
\usepackage{algorithm}
\usepackage{algorithmic}
\usepackage{array}
\usepackage{booktabs}
\usepackage{multirow}
\usepackage{tabularx}
\usepackage{colortbl}
\usepackage{graphicx}
\usepackage{textcomp}
\usepackage{stfloats}
\usepackage{url}
\usepackage{verbatim}

\usepackage{nicefrac}
\usepackage{microtype}
\usepackage{xcolor}
\usepackage{multicol}
\usepackage{newfloat}
\usepackage{listings}
\usepackage{tcolorbox}
\usepackage{lscape}
\usepackage{pifont}
\usepackage{wrapfig}
\usepackage{floatflt}
\usepackage{marvosym}
\usepackage{afterpage}
\usepackage{fontawesome}
\usepackage[normalem]{ulem}

\usepackage{subcaption}

\usepackage[numbers,sort&compress]{natbib}

\definecolor{cvprblue}{rgb}{0.21,0.49,0.74}
\definecolor{c1}{HTML}{F2C335}
\definecolor{c2}{HTML}{D9666F}
\definecolor{c3}{HTML}{FCEDC6}
\definecolor{c4}{HTML}{285D9B}
\definecolor{mygray}{HTML}{E6F0E8}
\definecolor{color1}{HTML}{ECF4F9}
\definecolor{color2}{HTML}{FFF1E0}
\definecolor{color3}{HTML}{F8F4F8}
\definecolor{Gray}{gray}{0.95}
\definecolor{LightBlue}{HTML}{F9FBF6}
\definecolor{vis1}{HTML}{C00000}
\definecolor{vis2}{HTML}{002060}
\definecolor{vis3}{HTML}{196B24}
\definecolor{mydeepgreen}{RGB}{0,100,0}
\definecolor{deepred}{RGB}{139,0,0}

\newcommand{\CC}[1]{\cellcolor{LightBlue}}
\newcommand{\RC}[1]{\rowcolor{LightBlue}}

\usepackage[breaklinks,colorlinks,allcolors=cvprblue]{hyperref}
\hypersetup{
    colorlinks=true,
    urlcolor=c2,
}

\title{VLMs are Good Teachers for Video Reasoning via Adaptive Test-Time Optimization}

\author{Junhao~Cheng, Liang~Hou, Tianxiong~Zhong, Xin~Tao, Pengfei~Wan, Kun~Gai, and Jing~Liao
\thanks{Junhao Cheng and Jing Liao are with City University of Hong Kong.}
\thanks{Liang Hou, Tianxiong Zhong, Xin Tao, Pengfei Wan, and Kun Gai are with Kling Team, Kuaishou Technology.}
\thanks{Corresponding author: Jing Liao.}
}

\begin{document}

\maketitle

\begin{abstract}
The emerging ``Reasoning with Video" paradigm utilizes Video Generation Models (VGMs) to generate temporally coherent visual trajectories to complete reasoning tasks. Although state-of-the-art VGMs excel at visual quality, they often struggle to understand and adhere to task-specific rules, leading to logical mistakes across diverse reasoning scenarios. Existing efforts employ Vision-Language Models (VLMs) as problem pre-solvers to produce or refine textual guidance for VGMs. However, textual descriptions fail to capture intricate spatiotemporal details, and VGMs often struggle to faithfully execute fine-grained or long-tail instructions even with a valid plan. While VLMs struggle as solvers, they possess strong perception capabilities to evaluate whether process constraints are satisfied and the final goal is achieved. Leveraging this strength, we introduce a paradigm shift that transitions the role of VLMs to ``teachers''. Specifically, a VLM teacher extracts task-specific rules to formulate differentiable rewards, guiding a VGM Reasoner via test-time optimization of a lightweight LoRA module. This strategy enables instance-specific adaptation at inference time and extends the reasoning capabilities beyond the VGM's intrinsic boundaries. Extensive experiments on VBVR-Bench (symbolic reasoning) and RULER-Bench (general reasoning) show that our method gains 16.7 points on average. It surpasses the VLM-as-Solver scheme (+0.4 points) and Best-of-N scaling (+2.2 points) by a large margin under comparable test-time computational cost. These findings reveal that integrating VLMs as test-time teachers offers a promising paradigm for achieving generalizable video reasoning. 

\noindent Project Page: {\url{https://VLM-as-Teacher.github.io/}}

\end{abstract}


\section{Introduction}
\label{sec: intro}

\begin{figure*}[!t]
	\centering
	\includegraphics[width=\textwidth]{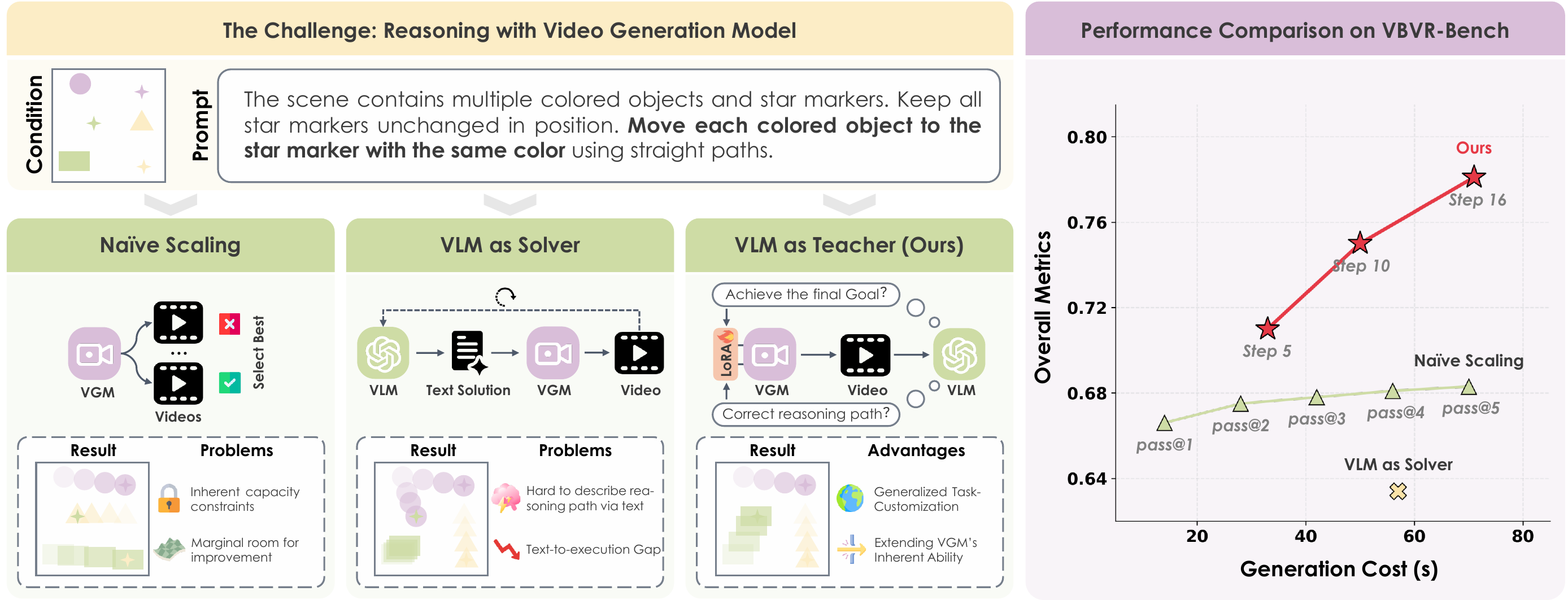}
\caption{\textbf{Vision-Language Models (VLMs) as Teachers Rather Than Solvers for Video Reasoning.} Unlike Test-time scaling with fixed Video Generation Models (VGMs) or VLM-as-Solver methods that rely on textual guidance, our VLM-as-Teacher paradigm supervises process constraints and final-goal achievement to guide a VGM Reasoner via online test-time optimization.}
\label{fig:teaser}
\end{figure*}

Recent advancements in Video
Generation Models (VGMs) demonstrate strong performance in synthesizing realistic and temporally coherent videos~\cite{openai2024sora, seedance2026seedance, wan2025wan}. Beyond content creation, several pioneering studies~\cite{tong2025think, guo2025mmecof} try to employ VGMs to solve logical reasoning tasks, forming an emerging research direction called ``Reasoning with Video''. By generating coherent visual trajectories, VGMs can address vision-centric reasoning challenges that are difficult to specify using language alone, such as the precise rotation of irregular objects. In certain tasks such as maze solving and puzzles, VGMs have been shown to match or even exceed the performance of state-of-the-art (SOTA) Vision-Language Models (VLMs) that rely primarily on textual reasoning chains~\cite{li2026thinking}. However, the optimization goal of VGMs is primarily visual fidelity~\cite{lipman2022flow,bishop2006pattern}, leading to the models' intrinsic limitations in performing logical reasoning and following task-specific rules. As a result, they often generate trajectories that are visually plausible but logically inconsistent with the goals.

To address the intrinsic limitations of VGMs, some efforts have explored test-time scaling (TTS) strategies, such as Best-of-N sampling or rejection-based schemes~\cite{newman2026video} in video reasoning. As illustrated in Fig.~\ref{fig:teaser}, these methods keep the VGM fixed and search among sampled videos. While effective at reducing stochastic errors, these approaches provide limited gains in video reasoning tasks. Systematic failures such as logically inconsistent trajectories and missed causal dependencies, cannot be easily corrected through repeated sampling because the model's inherent generative capacity constrains the solution space. Another line of work explores integrating VLMs as pre-solvers or planners to guide video reasoning~\cite{kim2026collabvr, chen2025tivibench}. As shown in Fig.~\ref{fig:teaser}, this ``VLM-as-Solver'' paradigm provides textual guidance for the VGM. However, reasoning via text alone remains challenging: linguistic prompts often fail to capture intricate spatiotemporal constraints, and even when a plan is detailed and logically sound, VGMs frequently struggle to faithfully execute fine-grained or long-tail instructions despite receiving a valid plan~\cite{cheng2025video}. 

Nevertheless, VLMs that struggle to construct executable visual solution trajectories are well suited to verifying whether a generated trajectory satisfies observable process constraints and reaches the intended final goal. For instance, even when a VLM cannot plan the exact steps for navigating a ball through a maze, it can evaluate whether the ball reaches the exit and whether its trajectory preserves the ball's identity and avoids crossing walls. Together, these conditions characterize successful task completion. Leveraging this strength, we uncover a new role for VLMs as ``teachers'', as shown in Fig.~\ref{fig:teaser}. In this paradigm, a VLM extracts task-specific rules and formulates them as differentiable rewards by proposing queries that assess whether intermediate steps adhere to constraints and whether the final state satisfies the intended goal. Unlike TTS, which keeps the VGM fixed and searches among sampled videos, these rewards guide a VGM Reasoner through test-time optimization (TTO), where instance-specific parameters are optimized under a task-specific objective during inference. Specifically, we optimize a lightweight LoRA module~\cite{hu2022lora} to adapt the VGM Reasoner to each reasoning instance, while using early video prediction, lightweight decoding, and loss-based early stopping to keep per-instance optimization efficient. By directly backpropagating differentiable feedback from the VLM, the VGM can refine its reasoning trajectories during inference, effectively aligning rule logic with visual execution and extending capabilities beyond its intrinsic limits.

Evaluations on symbolic (VBVR-Bench~\cite{wang2026very}) and general-purpose (RULER-Bench~\cite{he2025rulerbench}) video reasoning benchmarks show that the proposed method yields a 16.7-point average performance gain, comparing favorably against the VLM-as-Solver paradigm (+0.4 points) and Best-of-N scaling (+2.2 points) at comparable test-time cost, offering a promising paradigm to empower reasoning in video generation models.

We make the following contributions in this work:
\begin{itemize}
    \item We uncover a new VLM-as-Teacher paradigm for video reasoning, which fundamentally shifts the role of VLMs from text-based solvers to test-time supervisors that provide optimization signals for reasoning.
    \item We introduce a test-time online optimization approach for VGMs that adapts a VGM through differentiable VLM rewards, enabling reasoning capability beyond the model's intrinsic generative limits.
    \item We propose a task-adaptive reward synthesis strategy that automatically derives process and goal rewards from task descriptions, which together serve as sufficient conditions for successful reasoning task completion.
\end{itemize}

\section{Related Work}
\label{sec: related work}

\vspace{1mm} \noindent  \textbf{Reasoning with Video.} 
Since the emergence of diffusion models and transformer-based scaling~\cite{ho2020denois, peebles2023scala, zhang2024motiondiffuse}, video generation models have witnessed rapid proliferation. This includes closed-source pioneers such as Sora, Veo, and Seedance, as well as open-source counterparts like CogVideoX, HunyuanVideo, and Wan~\cite{openai2024sora, polyak2024movie, deepmind2025veo3, seedance2026seedance, gao2025seedance, seedance2025seedance, yang2024cogvid, kong2024hunyuan, wan2025wan}. While these models excel at synthesizing videos with high visual fidelity~\citep{peebles2023scala, yang2024cogvid, zheng2024opensora, huang2026vbenchpp}, recent research has further sought to optimize their alignment with physical laws and real-world dynamics~\cite{agarwal2025cosmos, zhang2025videorepa, wang2025wisa, chen2025towards, xue2025phyt2v, zhang2024physdreamer, liu2024physgen, gao2024flip, xie2025physanimator, montanaro2024motioncraft, wang2025prophy, yuan2025magictime}. Despite these advancements in visual and physical realism, they are not specifically optimized for rule-based relational, causal, or counterfactual reasoning.

To bridge this gap, the emerging ``Thinking with Frames'' paradigm re-conceptualizes video generation as a computational substrate for visual reasoning rather than mere synthesis~\cite{tong2025think, guo2025mmecof, liu2025genvire, wiedemer2025videozeroshot}. Preliminary studies on models like Veo-3 provide early evidence that large-scale pre-training can evoke non-trivial zero-shot perceptual and manipulation behaviors, enabling the solution of simple tasks without task-specific fine-tuning~\citep{wiedemer2025videozeroshot}. Drawing an analogy to the Chain-of-Thought (CoT) prompting in LLMs~\cite{wei2022chain}, recent works suggest that reasoning emerges through multi-step ``Chain-of-Frame" (CoF) diagnosis~\cite{guo2025mmecof, liu2025genvire, qi2026mme, li2026thinking}, where extended temporal sequences serve as explicit reasoning trajectories. Conversely, Wang et al.~\cite{wang2026demystifying} argue that reasoning processes are latent within the early stages of the denoising process, formulated as ``Chain-of-steps" (CoS) reasoning.
To quantify these capabilities, various benchmarks have been established to evaluate reasoning through synthetic puzzles such as maze solving and Sudoku~\cite{yang2025vrbench, cai2025mmgr}, as well as complex Text-Image-to-Video (TI2V) tasks~\cite{luo2025vreasonbench, chen2025tivibench, zhang2025ui2v}. Large-scale synthetic datasets now span five core dimensions, including perception, transformation, spatiality, abstraction, and knowledge, and encompass thousands of diverse tasks~\cite{wang2026very}. Beyond symbolic visual reasoning, benchmarks such as RULER-Bench~\cite{he2025rulerbench} and FAR~\cite{zhang2026far} further evaluate general-purpose video reasoning in open-ended scenarios. Despite the rapid development of benchmarks and diagnostic analyses, generalizable algorithmic solutions that bridge visual synthesis and logical rule adherence remain scarce.

\vspace{1mm} \noindent  \textbf{Test-Time Scaling for Video Reasoning.}
Test-time scaling has emerged as a powerful mechanism to enhance the performance of Large Language Models (LLMs)~\cite{snell2024scaling, brown2024monkeys} and diffusion models~\cite{ma2025inferencetime} by allocating additional compute during inference without modifying model parameters. Recent video-specific extensions~\cite{liu2025videot1, he2025evosearch, cong2025swift, li2026thinkinginframes, jang2026selfrefining} extend this concept to the temporal axis through frame-level tree searches, evolutionary sampling, and iterative self-refinement. Specifically for video reasoning, several approaches adapt Best-of-N scaling strategies; for instance, Wang et al.~\cite{wang2026demystifying} aggregate early denoising layers across different sampling seeds to produce optimal results, while EPBS~\cite{newman2026video} leverages the ``early commitment'' characteristic of video reasoning to accelerate the scaling process. 

However, these methods are fundamentally constrained by the inherent generative capacity of the base models. In complex reasoning tasks, failures are often systematic, such as logically flawed solution paths, skipped sub-goals, or physically inconsistent outcomes, rather than stochastic errors that can be mitigated through repeated sampling. Consequently, simply increasing test-time scaling through rejection sampling or ensemble methods yields limited gains. This motivates a different form of test-time computation: test-time optimization, which optimizes instance-specific variables or parameters under a test-time objective. In this work, we adopt TTO for video reasoning, allowing the VGM Reasoner to actively adapt toward rule-compliant visual trajectories.

\begin{figure*}[!t]
	\centering
	\includegraphics[width=\textwidth]{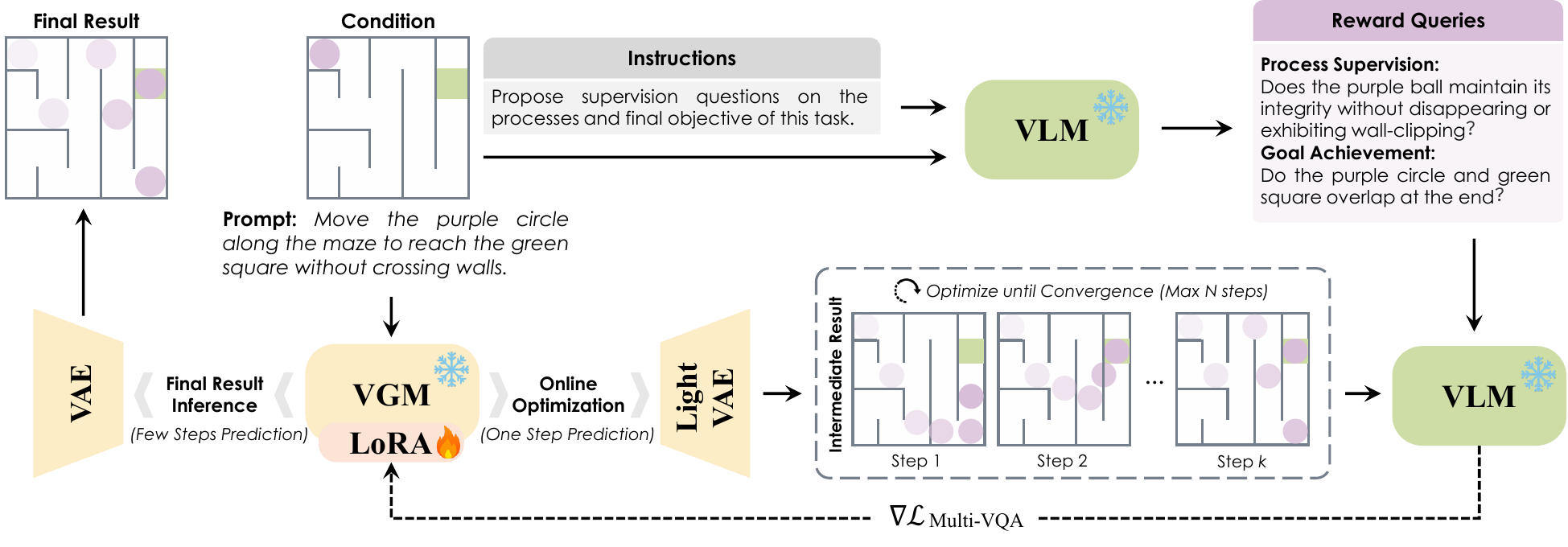}
\caption{Adaptive test-time optimization with a VLM Teacher. Given a rule-based video reasoning task, the VLM Teacher extracts task-specific process constraints and the final goal, and converts them into reward queries. During online optimization, an intermediate video prediction from the VGM is evaluated by the VLM Teacher. The resulting differentiable feedback updates a LoRA module. The optimized VGM then produces the final visual reasoning trajectory after the optimization loop ends.}
\label{fig: method}
\end{figure*}

\vspace{1mm} \noindent  \textbf{Integrating VLMs for Video Reasoning.}

Vision-Language Models (VLMs) possess formidable perceptual and reasoning capabilities, making them ideal candidates for enhancing reasoning tasks~\cite{zhang2024vision, li2025otter, liu2023cross, hui2023language, cheng2025videoHolmes,chen2025grpo}. Current LLM/VLM-guided generation paradigms typically cast the large model as a symbolic planner or a problem solver. These approaches, originating in the image domain~\cite{ke2024sld, yang2024rpg,xiao2026mindomni} and extending to video~\cite{lin2024videodirectorgpt, yang2025vlipp, xue2025phyt2v, wang2024videoagent, huang2025vchain, cheng2025video}, primarily optimize visual or physical attributes through text-based orchestration. 
Recent efforts have attempted to adapt this paradigm to video reasoning; for instance, VideoTPO~\cite{chen2025tivibench} uses LLM critiques to iteratively refine prompts, while CollabVR~\cite{kim2026collabvr} employs the VLM as a progressive planner and solver.  
However, these systems rely heavily on textual prompts, which often struggle to capture intricate spatiotemporal nuances. Furthermore, even with a logically sound plan, VGMs frequently fail to execute fine-grained or long-tail concepts due to the inherent gap between linguistic instructions and visual synthesis.
While VLMs struggle as solvers, they excel at evaluating generative processes. We therefore transition the role of a VLM from a ``solver'' to a ``teacher''. Specifically, a VLM Teacher formulates differentiable rewards from task-specific rules and guides a VGM through test-time optimization, bridging the gap between high-level logic and visual execution.

\section{Method}
\label{sec: Method}

\subsection{Task Formulation}

In this paper, we study rule-based video reasoning, where a VGM produces a temporally coherent visual trajectory (a video) that follows task-specific rules and achieves an intended goal. This setting covers symbolic visual reasoning tasks, such as spatial navigation, geometric manipulation, object arrangement, and sequential state transformation~\cite{li2026thinking,newman2026video}, as well as general-purpose scenarios, such as anomaly removal, object rotation, and hypothesis generation~\cite{wang2026very,he2025rulerbench}.

Formally, a reasoning instance is specified by a condition $\mathbf{c}=(\mathbf{p},\mathbf{x})$, where $\mathbf{p}$ denotes a textual instruction and $\mathbf{x}$ denotes an optional condition image. Given $\mathbf{c}$, a VGM $G_{\theta}$ generates a video as a visual reasoning trajectory:
\begin{equation}
    \mathbf{v}
    =
    G_{\theta}(\mathbf{c}; \epsilon)
    =
    \{v_1, v_2, \ldots, v_T\},
\end{equation}
where $\theta$ denotes the parameters of the VGM and $\epsilon$ denotes the sampling randomness. Following prior formulations~\cite{wang2026very,he2025rulerbench,guo2025mmecof}, successful task completion requires achieving the final goal while satisfying the process constraints. We denote the final-goal predicate by $g(\mathbf{v},\mathbf{c})$ and the set of process-constraint predicates by $\mathcal{R}(\mathbf{v},\mathbf{c})=\{r_m(\mathbf{v},\mathbf{c})\}_{m=1}^{M}$. Accordingly, task success is formulated as
\begin{equation}
    \operatorname{Succ}(\mathbf{v}, \mathbf{c})
    =
    \mathbb{I}
    \left[
    g(\mathbf{v}, \mathbf{c})=1
    \ \land\
    \bigwedge_{m=1}^{M} r_m(\mathbf{v}, \mathbf{c})=1
    \right].
\end{equation}
The central challenge is that the required rules vary across individual tasks and conditions. It is difficult for a general set of reward functions to characterize diverse task-specific constraints~\cite{zhu2026video}. To address this, we use a VLM Teacher to synthesize supervision queries for each case and directly guide the VGM via test-time optimization. 

\subsection{VLM-as-Teacher Framework}

Fig.~\ref{fig: method} illustrates the proposed VLM-as-Teacher framework, which consists of a VLM Teacher and a VGM Reasoner equipped with a lightweight LoRA module for test-time optimization. Rather than generating a textual solution trajectory, the VLM Teacher first identifies the requirements for successful task completion and then provides differentiable supervision to optimize the VGM Reasoner. This raises the challenge of how to convert the teacher's evaluative feedback into an effective optimization signal for video reasoning.

Recent pioneering studies have shown that VLM feedback can be formulated as a differentiable objective for generative models~\cite{luo2025dual,kumari2025learning,wang2026diffusion}. Luo et al.~\cite{luo2025dual} apply differentiable VLM rewards to image generation with manually specified queries, while others~\cite{kumari2025learning,wang2026diffusion} use VLM feedback to post-train generative models toward general visual quality rather than performing test-time optimization. Different from these works, we adapt differentiable VLM feedback to task-adaptive video reasoning. The VLM Teacher automatically derives goal-achievement and process-supervision queries from each reasoning condition, instead of relying on manually specified or task-agnostic queries. The resulting rewards are used to optimize the VGM Reasoner for the current test instance, rather than serving as a shared post-training objective. To make such video-level test-time optimization practical, we update only a lightweight LoRA module and evaluate an efficient first-step video prediction with a surrogate VAE decoder. In this way, VLM feedback directly supervises rule satisfaction, final-goal achievement, and reasoning-trajectory validity with manageable optimization cost. The overall procedure is summarized in Algorithm~\ref{alg:vlm_teacher}.

\vspace{1mm} \noindent  \textbf{Task-Adaptive Supervision Synthesis.}
Given a task condition $\mathbf{c}$, the VLM Teacher first analyzes the textual instruction and the optional visual context to identify the success requirements of the task. It then formulates these requirements as binary reward queries. Specifically, the teacher synthesizes one goal achievement query $q_{\mathrm{goal}}(\mathbf{c})$ and $M$ process supervision queries $\{q_{\mathrm{proc}}^{m}(\mathbf{c})\}_{m=1}^{M}$, where typically $1 \leq M \leq 3$. The resulting query set is defined as
\begin{equation}
    \mathcal{Q}(\mathbf{c})
    =
    \left\{ q_{\mathrm{goal}}(\mathbf{c}) \right\}
    \cup
    \left\{ q_{\mathrm{proc}}^{m}(\mathbf{c}) \right\}_{m=1}^{M}.
\end{equation}
The process supervision queries evaluate whether the generated trajectory follows the task-specific rules, such as object integrity, valid motion, temporal continuity, collision constraints, or state consistency. The goal achievement query evaluates whether the final state satisfies the intended objective. For example, in the maze navigation task shown in Fig.~\ref{fig: method}, the teacher generates process queries that examine whether the purple ball remains intact and avoids crossing walls, together with a goal query that examines whether the ball reaches the green target region.

\begin{algorithm}[t]
\caption{Test-time Optimization with a VLM Teacher}
\label{alg:vlm_teacher}
\begin{algorithmic}[1]
\REQUIRE Task condition $\mathbf{c}=(\mathbf{p},\mathbf{x})$; VGM Reasoner $G_{\theta,\phi}$; VLM Teacher $P_{\psi}$; lightweight surrogate decoder $D_{\mathrm{lite}}$; maximum optimization steps $N$; loss threshold $\tau_{\mathcal{L}}$; sampled frame number $K$
\ENSURE Final visual reasoning trajectory $\mathbf{v}^{*}$
\STATE $\mathcal{Q}(\mathbf{c})
\leftarrow
\textsc{SynthesizeQueries}
\left(
P_{\psi},\mathbf{c}
\right)$
\STATE Initialize LoRA parameters $\phi_{0}$; set $\phi^{*} \leftarrow \phi_{0}$
\STATE Sample initial pure-noise latent $z_{1}=\epsilon$
\FOR{$n=0,\ldots,N-1$}
    \STATE $\hat{z}_{0}^{(n)}
    \leftarrow
    z_{1}
    -
    u_{\theta,\phi_n}
    \left(
    z_{1},1,\mathbf{c}
    \right)$
    \STATE $\tilde{\mathbf{v}}^{(n)}
    \leftarrow
    \textsc{Sample}_{K}
    \left(
    D_{\mathrm{lite}}
    \left(
    \hat{z}_{0}^{(n)}
    \right)
    \right)$
    \STATE $\mathcal{L}_{\mathrm{Multi\text{-}VQA}}^{(n)}
    \leftarrow
    \textsc{VLM-Loss}
    \left(
    P_{\psi};
    \tilde{\mathbf{v}}^{(n)},
    \mathcal{Q}(\mathbf{c})
    \right)$
    \IF{$\mathcal{L}_{\mathrm{Multi\text{-}VQA}}^{(n)}
    \leq
    \tau_{\mathcal{L}}$}
        \STATE $\phi^{*} \leftarrow \phi_n$
        \STATE \textbf{break}
    \ENDIF
    \STATE $\phi_{n+1}
    \leftarrow
    \phi_n
    -
    \eta
    \nabla_{\phi_n}
    \mathcal{L}_{\mathrm{Multi\text{-}VQA}}^{(n)}$
    \STATE $\phi^{*} \leftarrow \phi_{n+1}$
\ENDFOR
\STATE $\mathbf{v}^{*}
\leftarrow
G_{\theta,\phi^{*}}(\mathbf{c};\epsilon)$
\COMMENT{Decode with the standard VAE}
\RETURN $\mathbf{v}^{*}$
\end{algorithmic}
\end{algorithm}

All reward queries are phrased positively, i.e., a ``Yes'' response indicates satisfaction of the corresponding requirement. This formulation provides a unified reward interface for heterogeneous rule-based reasoning tasks without manually defining reward functions for individual task categories. Moreover, the two types of supervision are complementary: the goal achievement query alone does not prevent invalid intermediate trajectories, while the process supervision queries alone do not ensure successful task completion.

\vspace{1mm} \noindent  \textbf{Online Optimization Process.}
With the reward queries, we next utilize the VLM Teacher to guide the reasoning trajectory of the VGM Reasoner. We apply differentiable VLM supervision to test-time optimization of a VGM Reasoner, enabling task-specific optimization for each rule-based video reasoning instance.

For each reasoning instance, the pretrained VGM backbone and the VLM Teacher remain frozen, and only a lightweight LoRA module is optimized. Let $\phi_n$ denote the LoRA parameters at the $n$-th optimization step, and let $\tilde{\mathbf{v}}^{(n)}$ denote the intermediate video result evaluated by the VLM Teacher. Following the differentiable VQA formulation, the VLM Teacher evaluates each video-query pair by predicting a target answer sequence. Since all synthesized reward queries are positively phrased, the target answer for every query is the response ``Yes''. We denote the tokenized target answer by
$S_a^{+}=\operatorname{Tok}(\texttt{Yes})=\{a_{\ell}^{+}\}_{\ell=1}^{L}$, where $L$ is the number of tokens. For each reward query $q \in \mathcal{Q}(\mathbf{c})$, we define the corresponding VQA loss as
\begin{equation}
    \mathcal{L}_{\mathrm{VQA}}
    \left(
    \tilde{\mathbf{v}}^{(n)},q
    \right)
    =
    -
    \sum_{\ell=1}^{L}
    \log
    P_{\psi}
    \left(
    a_{\ell}^{+}
    \mid
    \tilde{\mathbf{v}}^{(n)}, q, a_{<\ell}^{+}
    \right),
\end{equation}
where $P_{\psi}$ denotes the frozen VLM Teacher. Unlike visual instruction tuning, which optimizes the parameters of the VLM, the proposed objective propagates gradients through the visual prediction to optimize the LoRA parameters of the VGM Reasoner. Based on the synthesized query set, the complete objective consists of one goal achievement term and $M$ process supervision terms:
\begin{equation}
\begin{split}
    \mathcal{L}_{\mathrm{Multi\text{-}VQA}}^{(n)}
    =
    &
    \lambda
    \mathcal{L}_{\mathrm{VQA}}
    \left(
    \tilde{\mathbf{v}}^{(n)},
    q_{\mathrm{goal}}(\mathbf{c})
    \right)
    \\
    &
    +
    \frac{1-\lambda}{M}
    \sum_{m=1}^{M}
    \mathcal{L}_{\mathrm{VQA}}
    \left(
    \tilde{\mathbf{v}}^{(n)},
    q_{\mathrm{proc}}^{m}(\mathbf{c})
    \right),
\end{split}
\label{eq:multi_vqa}
\end{equation}
where $\lambda$ is a balance factor. The LoRA parameters are then updated by
\begin{equation}
    \phi_{n+1}
    =
    \phi_n
    -
    \eta
    \nabla_{\phi_n}
    \mathcal{L}_{\mathrm{Multi\text{-}VQA}}^{(n)},
\end{equation}
with learning rate $\eta$.

\vspace{1mm} \noindent  \textbf{Efficient Adaptation.}
Applying differentiable VLM supervision to video generation is computationally demanding, since a straightforward implementation requires repeated multi-step denoising, decoding with a heavy video VAE~\cite{wan2025wan}, and VLM evaluation during optimization~\cite{wang2026diffusion}. We introduce three designs to make the online optimization practical.

First, we replace the standard VAE with a lightweight surrogate decoder~\cite{lightx2v} during online optimization. This substantially reduces the memory and computation overhead of differentiable video decoding at the cost of moderate visual quality degradation. Experiments at Section 4.3 show that such degradation has a negligible effect on the VLM Teacher's evaluation accuracy. After optimization, the final visual reasoning trajectory is generated by the adapted VGM Reasoner and decoded using the standard VAE.

Second, we distill the VGM Reasoner into a four-step generator using~\cite{yin2024improved}, and update only its first-step clean-latent prediction during online optimization. Let $z_{1}=\epsilon$ denote the initial pure-noise latent, and let $u_{\theta,\phi_n}$ denote the velocity predicted by the adapted VGM Reasoner.
We obtain the one-step clean-latent prediction by applying the full sampling interval to this velocity prediction:
\begin{equation}
    \hat{z}_{0}^{(n)}
    =
    z_{1}
    -
    u_{\theta,\phi_n}
    \left(
    z_{1}, 1, \mathbf{c}
    \right),
    \qquad z_{1}=\epsilon.
\end{equation}
Recent analysis indicates that the high-level reasoning behavior of video generation models emerges in early denoising steps~\cite{wang2026demystifying}. In addition, we observe that the first-step prediction of a few-step Reasoner already provides a visually perceptible approximation of the reasoning trajectory. Therefore, the VLM Teacher can evaluate the reasoning behavior without repeatedly completing the full denoising process. We then decode and uniformly sample $K$ frames from the decoded first-step prediction as the input for VLM evaluation. Since the lightweight decoding and frame sampling operations preserve the computation graph, gradients from the VLM Teacher can be propagated through $\tilde{\mathbf{v}}^{(n)}$ to the LoRA parameters $\phi_n$.

Third, we employ loss-based early stopping to avoid unnecessary optimization steps. Since $\mathcal{L}_{\mathrm{Multi\text{-}VQA}}^{(n)}$ is defined by the negative log-likelihood of the positive answer ``Yes'' over the goal achievement query and all process supervision queries, a lower loss indicates that the VLM Teacher assigns higher confidence to the satisfaction of the task requirements. Online optimization terminates when
$\mathcal{L}_{\mathrm{Multi\text{-}VQA}}^{(n)} \leq \tau_{\mathcal{L}}$
or when the maximum number of optimization steps $N$ is reached, where $\tau_{\mathcal{L}}$ denotes the predefined loss threshold. The resulting LoRA module is then used by the VGM Reasoner to generate the final visual reasoning trajectory.

\begin{table*}[!t]
\centering
\caption{Benchmarking results on VBVR-Bench. Higher is better. \textbf{Cost} stands for average total inference generation seconds per sample. \textbf{Bold} stands for best in group; \underline{Underlined} stands for second best in group. Tasks include Abstraction (Abs.), Knowledge (Know.), Perception (Perc.), Spatiality (Spat.), and Transformation (Trans.).}
\label{tab:benchmarking-results}
\small
\setlength{\tabcolsep}{4pt}
\resizebox{\textwidth}{!}{
\begin{tabular}{l | c | c | c | c c c c c | c | c c c c c}
\toprule
\multirow{2}{*}{\textbf{Models}} & \multirow{2}{*}{\textbf{Cost (s)}} & \multirow{2}{*}{\textbf{Overall}} & \multicolumn{6}{c|}{\textbf{In-Domain by Category}} & \multicolumn{6}{c}{\textbf{Out-of-Domain by Category}} \\
\cmidrule(lr){4-9} \cmidrule(lr){10-15}
& & & \textbf{Avg.} & \textbf{Abst.} & \textbf{Know.} & \textbf{Perc.} & \textbf{Spat.} & \textbf{Trans.} & \textbf{Avg.} & \textbf{Abst.} & \textbf{Know.} & \textbf{Perc.} & \textbf{Spat.} & \textbf{Trans.} \\
\midrule
\multicolumn{15}{>{\columncolor{color3}}c}{\textbf{Closed-source Models}} \\
Sora 2 & - & \textbf{0.546} & \textbf{0.569} & \underline{0.602} & \underline{0.477} & \textbf{0.581} & \textbf{0.572} & \textbf{0.597} & \textbf{0.523} & \underline{0.546} & \textbf{0.472} & \textbf{0.525} & \textbf{0.462} & \textbf{0.546} \\
Kling 2.6 & - & 0.369 & 0.408 & 0.465 & 0.323 & 0.375 & 0.347 & \underline{0.519} & 0.330 & 0.528 & 0.135 & 0.272 & 0.356 & 0.359 \\
Veo 3.1 & - & \underline{0.480} & \underline{0.531} & \textbf{0.611} & \textbf{0.503} & \underline{0.520} & \underline{0.444} & 0.510 & \underline{0.429} & \textbf{0.577} & \underline{0.277} & \underline{0.420} & \underline{0.441} & \underline{0.404} \\
\midrule
\multicolumn{15}{>{\columncolor{color3}}c}{\textbf{Open-source Models}} \\
VBVR-Wan2.2-14B & 160 & 0.682 & \underline{0.763} & \underline{0.733} & 0.713 & \underline{0.795} & \underline{0.776} & \textbf{0.827} & 0.601 & \underline{0.732} & 0.596 & 0.542 & 0.628 & \underline{0.600} \\
VBVR-Wan2.2-5B & 87 & 0.676 & 0.713 & 0.675 & 0.722 & 0.715 & 0.733 & 0.715 & 0.639 & 0.711 & 0.618 & 0.642 & 0.678 & 0.548 \\
\ \  + Pass@2 & 174 & 0.690 & 0.729 & 0.686 & 0.749 & 0.727 & 0.751 & 0.727 & 0.650 & 0.718 & 0.659 & 0.647 & 0.680 & 0.559 \\
\ \  + Pass@3 & 261 & 0.693 & 0.733 & 0.693 & 0.751 & 0.728 & 0.753 & 0.736 & 0.652 & 0.720 & 0.660 & 0.650 & 0.682 & 0.560 \\
\ \  + Pass@4 & 348 & 0.700 & 0.740 & 0.713 & 0.757 & 0.729 & 0.756 & 0.740 & 0.660 & 0.723 & 0.661 & \underline{0.665} & 0.695 & 0.563 \\
\ \  + Pass@5 & 435 & \underline{0.701} & 0.741 & 0.714 & \underline{0.762} & 0.730 & 0.757 & 0.741 & \underline{0.661} & 0.725 & \underline{0.664} & \underline{0.665} & 0.695 & 0.564 \\
\ \ + VideoTPO & 276 & 0.663 & 0.697 & 0.654 & 0.701 & 0.698 & 0.724 & 0.708 & 0.629 & 0.703 & 0.604 & 0.631 & 0.669 & 0.538 \\
VBVR-Wan2.2-5B-Distilled & 14 & 0.666 & 0.692 & 0.638 & 0.709 & 0.661 & 0.732 & 0.712 & 0.640 & 0.688 & 0.603 & 0.651 & 0.693 & 0.565 \\
\ \  + Pass@2 & 28 & 0.675 & 0.702 & 0.653 & 0.717 & 0.674 & 0.743 & 0.718 & 0.647 & 0.701 & 0.603 & 0.651 & 0.707 & 0.575 \\
\ \  + Pass@3 & 42 & 0.678 & 0.707 & 0.659 & 0.719 & 0.679 & 0.748 & 0.722 & 0.650 & 0.702 & 0.603 & 0.651 & 0.720 & 0.576 \\
\ \  + Pass@4 & 56 & 0.681 & 0.711 & 0.673 & 0.722 & 0.680 & 0.748 & 0.726 & 0.652 & 0.703 & 0.603 & 0.651 & 0.727 & 0.581 \\
\ \  + Pass@5 & 70 & 0.683 & 0.712 & 0.676 & 0.722 & 0.680 & 0.751 & 0.726 & 0.653 & 0.709 & 0.603 & 0.651 & \underline{0.728} & 0.582 \\
\ \ + VideoTPO & 57 & 0.634 & 0.671 & 0.624 & 0.687 & 0.643 & 0.712 & 0.689 & 0.597 & 0.652 & 0.584 & 0.613 & 0.641 & 0.495 \\
\rowcolor[HTML]{F4F8EE}  \ \ \textbf{+ Ours} & 69 & \textbf{0.781} & \textbf{0.803} & \textbf{0.806} & \textbf{0.920} & \textbf{0.837} & \textbf{0.820} & \underline{0.787} & \textbf{0.759} & \textbf{0.873} & \textbf{0.765} & \textbf{0.759} & \textbf{0.818} & \textbf{0.639} \\
\bottomrule
\end{tabular}
}
\end{table*}

\begin{table*}[!t]
\centering
\caption{Benchmarking results on RULER-Bench. Higher is better. \textbf{Cost} stands for average total generation seconds per sample. \textbf{Bold} denotes the best result in each group, and \underline{underlined} denotes the second best. Tasks include Transportation (Tra.), Sports (Spo.), Social (Soc.), Safety (Saf.), Festival (Fes.), Dress (Dre.), Food (Foo.), Emotion (Emo.), Chemistry (Che.), Physics (Phy.), Biology (Bio.), Earth Science (Ear.), Mathematics (Mat.), Medicine (Med.), Life (Lif.), Subjective (Sub.), Objective (Obj.), Idiom (Idi.), Metaphor (Met.), Definition (Def.), Anomaly (Ano.), Color (Col.), Count (Cou.), Direction (Dir.), Position (Pos.), Shape (Sha.), Size (Siz.), Style (Sty.), Viewpoint (Vie.), and Motion (Mot.).}

\label{tab:ruler-task-results}
\tiny
\setlength{\tabcolsep}{1.2pt}
\resizebox{\textwidth}{!}{
\begin{tabular}{l | c | c | cccccccc | ccccccc | cc | ccc | cccccccccc}
\toprule
\multirow{2}{*}{\textbf{Models}} 
& \multirow{2}{*}{\textbf{Cost (s)}} 
& \multirow{2}{*}{\textbf{Avg.}} 
& \multicolumn{8}{c|}{\textbf{Humanity}} 
& \multicolumn{7}{c|}{\textbf{Science}} 
& \multicolumn{2}{c|}{\textbf{Hypothesis}} 
& \multicolumn{3}{c|}{\textbf{Semantics}} 
& \multicolumn{10}{c}{\textbf{Vision}} \\
\cmidrule(lr){4-11} 
\cmidrule(lr){12-18} 
\cmidrule(lr){19-20} 
\cmidrule(lr){21-23} 
\cmidrule(lr){24-33}
& & 
& \text{Tra.} & \text{Spo.} & \text{Soc.} & \text{Saf.} & \text{Fes.} & \text{Dre.} & \text{Foo.} & \text{Emo.} 
& \text{Che.} & \text{Phy.} & \text{Bio.} & \text{Ear.} & \text{Mat.} & \text{Med.} & \text{Lif.} 
& \text{Sub.} & \text{Obj.} 
& \text{Idi.} & \text{Met.} & \text{Def.} 
& \text{Ano.} & \text{Col.} & \text{Cou.} & \text{Dir.} & \text{Pos.} & \text{Sha.} & \text{Siz.} & \text{Sty.} & \text{Vie.} & \text{Mot.} \\
\midrule

\multicolumn{33}{>{\columncolor{color3}}c}{\textbf{Closed-source Models}} \\

Veo 3.1 
& - 
& \underline{65.0} 
& \textbf{80.0} & 78.1 & 74.9 & 82.8 & 80.8 & 90.9 & \underline{90.9} & \textbf{69.2}
& 81.1 & \underline{79.1} & \underline{83.3} & \underline{57.3} & 61.5 & 63.4 & \underline{74.5}
& \underline{74.1} & 81.3
& \underline{76.5} & \underline{81.4} & \textbf{87.3}
& \textbf{42.2} & \textbf{63.9} & 20.1 & \underline{32.5} & 37.5 & \underline{46.9} & \textbf{47.9} & \textbf{47.5} & \textbf{51.3} & \underline{55.3} \\

\ \ + PE 
& - 
& \textbf{66.2} 
& \underline{78.5} & \textbf{82.4} & 75.1 & \underline{85.5} & \textbf{90.0} & \textbf{94.2} & \textbf{93.4} & 59.5
& \textbf{86.6} & 74.5 & 81.2 & \textbf{64.4} & 61.3 & \underline{68.1} & \textbf{83.7}
& \textbf{75.4} & \underline{81.9}
& \textbf{83.8} & 79.7 & \underline{86.6}
& \underline{38.4} & \underline{62.5} & 26.4 & \textbf{36.9} & 28.1 & \textbf{50.8} & \underline{46.4} & \textbf{47.5} & \textbf{51.3} & \textbf{57.9} \\

Sora 2 
& - 
& 59.8 
& 71.6 & 73.5 & \textbf{77.8} & 80.1 & 84.4 & 89.8 & 88.9 & \underline{63.6}
& \underline{85.8} & 76.4 & \textbf{84.1} & 52.9 & \underline{70.1} & 64.6 & 65.9
& 62.7 & 73.3
& 72.1 & 70.4 & 80.1
& 32.8 & 22.9 & \underline{34.0} & 30.0 & \underline{41.9} & 39.6 & 38.0 & 43.8 & 31.3 & 41.8 \\

\ \ + PE 
& - 
& 62.9 
& 69.4 & \underline{79.7} & \underline{75.2} & \textbf{87.9} & \underline{89.7} & \underline{93.7} & 85.7 & 55.6
& 81.6 & \textbf{82.3} & 76.4 & 51.1 & \textbf{72.9} & \textbf{80.6} & 73.5
& 73.0 & \textbf{84.3}
& 75.1 & \textbf{82.1} & 79.0
& 35.6 & 38.2 & \textbf{38.2} & 28.8 & \textbf{42.5} & 40.3 & 31.3 & \underline{46.9} & \underline{32.5} & 53.7 \\

\midrule
\multicolumn{33}{>{\columncolor{color3}}c}{\textbf{Open-source Models}} \\

Wan2.2-14B 
& 212 
& 49.4 
& 50.7 & 59.7 & \underline{51.7} & 59.4 & 68.0 & 85.7 & 61.0 & 40.1
& 51.6 & \underline{54.2} & 52.4 & 48.9 & 43.3 & 47.8 & 55.6
& 44.9 & 61.7
& 57.8 & 70.5 & \underline{64.0}
& 28.4 & 33.3 & 38.9 & 44.4 & 36.9 & 37.5 & 41.7 & 32.5 & 38.1 & 50.2 \\

Wan2.2-5B 
& 98 
& 46.7 
& 48.6 & 57.5 & 47.9 & 56.5 & 65.2 & 81.7 & 59.1 & 37.7
& 49.2 & 50.7 & 50.2 & 46.0 & 41.5 & 44.9 & 52.5
& 42.2 & 58.2
& 53.7 & 66.7 & 60.7
& 26.7 & 31.2 & 35.7 & 40.7 & 34.7 & 35.2 & 38.7 & 30.2 & 35.7 & 48.2 \\

\ \ + Pass@5 
& 490 
& 49.6 
& \underline{51.1} & 59.5 & 50.4 & 58.7 & 66.0 & 83.3 & 60.6 & \underline{41.7}
& 51.6 & 53.7 & \underline{54.2} & 48.5 & \underline{44.3} & 45.7 & 53.3
& 42.6 & 58.7
& 54.4 & 68.9 & 63.2
& \underline{32.5} & \underline{36.2} & 37.7 & \underline{45.9} & \underline{40.5} & 40.2 & \underline{44.5} & 35.2 & \underline{41.2} & 50.2 \\

\ \ + PE 
& 101 
& 48.6 
& 48.1 & 60.0 & 47.4 & 59.5 & 70.2 & 84.5 & 62.6 & 34.2
& 51.2 & 51.7 & 48.7 & 48.5 & 42.7 & 50.4 & 57.0
& 47.2 & 63.7
& 57.7 & 69.2 & 60.2
& 27.7 & 32.7 & 39.2 & 42.2 & 33.2 & 37.2 & 36.9 & 31.7 & 37.7 & 52.2 \\

\ \ + VideoTPO 
& 311 
& \underline{50.6} 
& 49.4 & \underline{62.0} & 48.6 & \underline{61.7} & \textbf{73.0} & \underline{86.6} & \underline{64.9} & 36.2
& \underline{53.3} & 53.5 & 49.9 & \underline{50.5} & 44.2 & \textbf{53.5} & \textbf{59.8}
& \textbf{50.6} & \textbf{67.1}
& \underline{60.5} & \underline{71.4} & 61.3
& 29.1 & 34.4 & 41.4 & 43.8 & 34.3 & 38.9 & 38.1 & 33.4 & 39.6 & 54.7 \\

Wan2.2-5B-Distilled 
& 18 
& 46.4 
& 47.7 & 57.0 & 47.2 & 56.0 & 64.6 & 81.0 & 58.5 & 37.3
& 48.3 & 50.0 & 49.6 & 45.3 & 41.2 & 44.0 & 52.1
& 41.7 & 57.7
& 53.3 & 66.3 & 60.2
& 26.7 & 31.5 & 36.0 & 40.8 & 35.0 & 35.6 & 39.0 & 30.6 & 35.9 & 48.8 \\

\ \ + Pass@5 
& 90 
& 49.1 
& 49.9 & 58.8 & 49.4 & 58.0 & 65.3 & 82.4 & 59.9 & 41.1
& 50.5 & 52.8 & 53.4 & 47.6 & 43.8 & 44.7 & 52.8
& 42.0 & 58.1
& 53.9 & 68.3 & 62.5
& 32.1 & \underline{36.2} & 37.8 & 45.6 & 40.4 & \underline{40.3} & 44.4 & \underline{35.3} & 41.0 & 50.6 \\

\ \ + PE 
& 20 
& 48.3 
& 47.2 & 59.5 & 46.7 & 59.0 & 69.6 & 83.8 & 62.0 & 33.8
& 50.3 & 51.0 & 48.1 & 47.8 & 42.4 & 49.5 & 56.6
& 46.7 & 63.2
& 57.3 & 68.8 & 59.7
& 27.7 & 33.0 & 39.5 & 42.3 & 33.5 & 37.6 & 37.2 & 32.1 & 37.9 & 52.8 \\

\ \ + VideoTPO 
& 71 
& 50.3 
& 48.5 & 61.5 & 47.9 & 61.2 & 72.4 & 85.9 & 64.3 & 35.8
& 52.4 & 52.8 & 49.3 & 49.8 & 43.9 & 52.6 & 59.4
& \underline{50.1} & 66.6
& 60.1 & 71.0 & 60.8
& 29.1 & 34.7 & \underline{41.7} & 43.9 & 34.6 & 39.3 & 38.4 & 33.8 & 39.8 & \underline{55.3} \\

\rowcolor[HTML]{F4F8EE}
\ \ \textbf{+ Ours} 
& 88 
& \textbf{68.2} 
& \textbf{78.6} & \textbf{79.8} & \textbf{75.3} & \textbf{85.6} & \underline{72.8} & \textbf{93.8} & \textbf{91.0} & \textbf{63.7}
& \textbf{85.9} & \textbf{79.2} & \textbf{83.4} & \textbf{57.4} & \textbf{70.2} & \underline{53.0} & \underline{59.6}
& 49.5 & \underline{66.9}
& \textbf{76.6} & \textbf{81.5} & \textbf{86.7}
& \textbf{65.0} & \textbf{66.1} & \textbf{61.3} & \textbf{64.8} & \textbf{46.9} & \textbf{54.0} & \textbf{51.3} & \textbf{51.6} & \textbf{56.0} & \textbf{69.1} \\

\bottomrule
\end{tabular}
}
\end{table*}

\section{Experiments}
\label{sec: exp}

\begin{figure*}[!t]
	\centering
	\includegraphics[width=\textwidth]{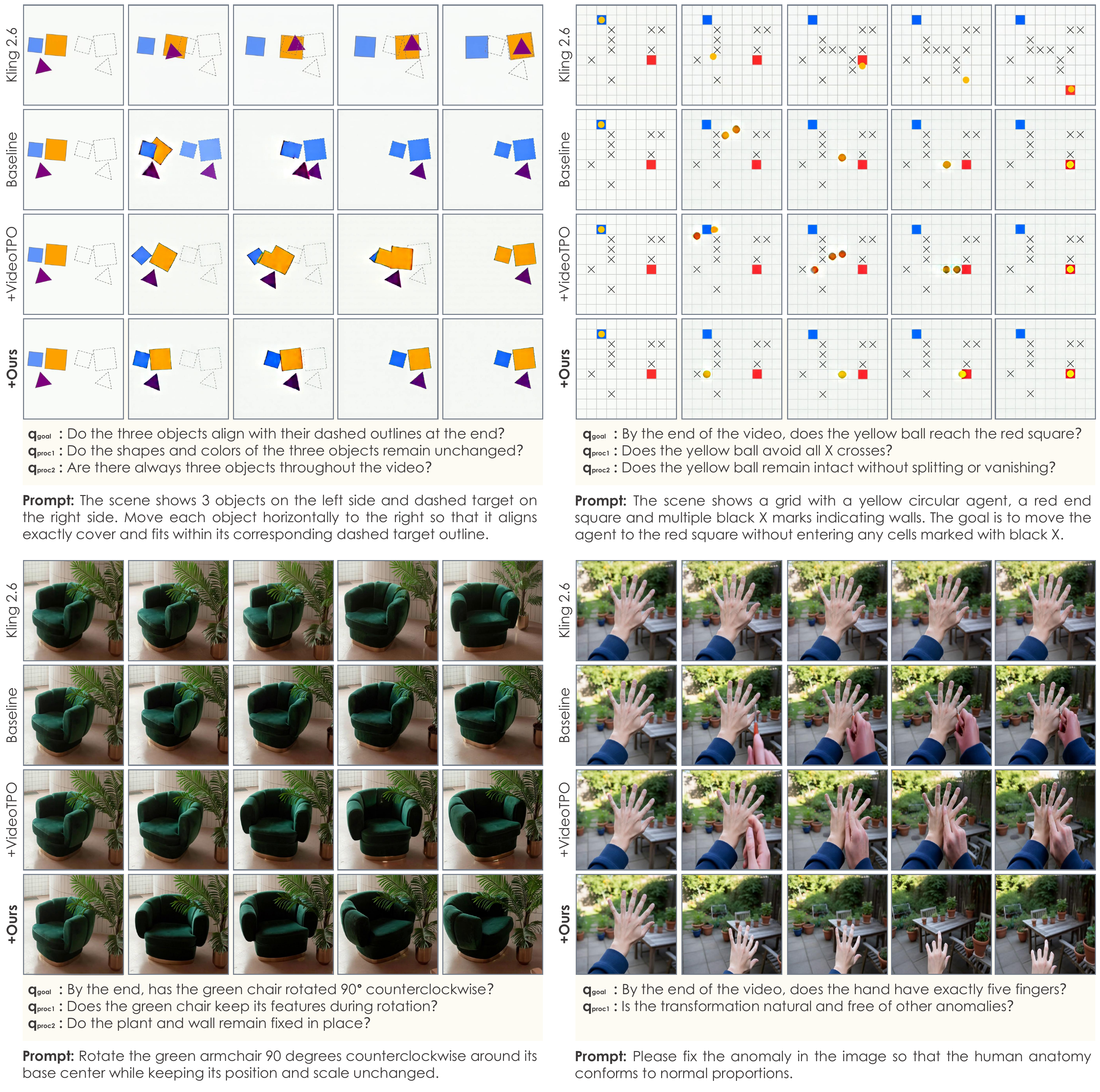}
\caption{Qualitative comparisons on symbolic and general-purpose video reasoning examples. Baseline stands for the step-distilled Wan2.2-5B model~\cite{wan2025wan}. $q_{\mathrm{goal}}$ and $q_{\mathrm{proc}}$ are representative supervision queries synthesized by the VLM Teacher. The proposed method satisfies both the final goal and the process constraints, leading to accurate reasoning results.}
\label{fig:main_vis}
\end{figure*}

\subsection{Experimental Setup}

\vspace{1mm} \noindent  \textbf{Benchmarks and Metrics.}
We evaluate the proposed method on two complementary video reasoning benchmarks. 
VBVR-Bench~\cite{wang2026very} focuses on symbolic visual reasoning tasks across five capability categories: abstraction, knowledge, perception, spatiality, and transformation. 
RULER-Bench~\cite{he2025rulerbench} contains general-purpose reasoning scenarios spanning six rule categories: humanity, science, hypothesis, semantics, vision, and game. 
Since the game tasks in RULER-Bench substantially overlap with the symbolic reasoning scenarios evaluated in VBVR-Bench, we exclude this category and evaluate the remaining 30 tasks from the other five categories. For VBVR-Bench, we report the overall score together with the in-domain (ID) and out-of-domain (OOD) averages. Since its tasks have verifiable outcomes, VBVR-Bench evaluates generated videos using task-specific rule-based detection scorers that measure spatial accuracy, trajectory correctness, temporal consistency, and logical validity. 
For RULER-Bench, we report the average score over the 30 evaluated task categories. 
Following its official protocol, each generated video is evaluated using checklist questions under four dimensions: instruction following, visual consistency, visual fidelity, and rule coherence. 
The checklist responses are scored by GPT-o3~\cite{openai2025o3}, following the evaluator adopted in the benchmark. For both benchmarks, we follow the officially released metrics and evaluation protocols to ensure fair comparison. We additionally report the average total generation time per sample for efficiency comparison.

\vspace{1mm} \noindent  \textbf{Compared Methods.}
We compare the proposed method with SOTA closed-source and open-source VGMs, including Sora~2~\cite{openai2024sora}, Kling~2.6~\cite{kuaishou2025kling26}, Veo~3.1~\cite{deepmind2025veo3}, and Wan2.2~\cite{wan2025wan}. 
Based on these generators, we compare three types of test-time reasoning strategies. 
\textbf{Pass@N} performs sampling-based test-time scaling by generating $N$ candidates with different initial noises and selecting the best result according to the evaluation criterion. 
\textbf{PE (Prompt Engineering)} and \textbf{VideoTPO} represent the ``VLM-as-Solver'' paradigm, where a VLM improves video generation through textual task specification. Specifically, \textbf{PE} uses a VLM to interpret the reasoning task and rewrite the initial prompt before video generation, while \textbf{VideoTPO} further observes generated results and iteratively refines the prompt through VLM feedback. 

\vspace{1mm} \noindent  \textbf{Implementation Details.}
Unless otherwise specified, we use a step-distilled Wan2.2-5B as our VGM Reasoner and Qwen3-VL-4B~\cite{bai2025qwen3} as the VLM Teacher. The VGM Reasoner is distilled into a four-step generator following DMD2~\cite{yin2024improved}. For VBVR-Bench, following the official setting~\cite{wang2026very}, we first perform domain-adaptive supervised fine-tuning on its 30K training instances for all open-source baselines.

During online optimization, only the LoRA parameters are updated. The first-step clean-latent prediction is decoded using the lightweight surrogate decoder from LightX2V~\cite{lightx2v}. We uniformly sample $K=24$ frames for VLM evaluation and set the maximum number of online optimization steps to $N=40$. The LoRA rank is set to $16$, the learning rate is $5\times10^{-5}$, and the loss balance factor is set to $\lambda=0.5$. We use a loss threshold of $\tau_{\mathcal{L}}=0.22$ for early stopping, which approximately corresponds to an overall VLM confidence of $0.8$ for answering ``Yes'' to the reward queries. After online optimization, the final video is generated by the optimized VGM Reasoner and decoded using the standard VAE. All compared open-source methods generate 89-frame videos under the same evaluation setting.

\subsection{Comparison with SOTA Methods}

\vspace{1mm} \noindent  \textbf{Quantitative Comparisons.} 
Tables~\ref{tab:benchmarking-results} and~\ref{tab:ruler-task-results} report the quantitative comparisons on VBVR-Bench and RULER-Bench, respectively. 
Notably, step distillation largely preserves the reasoning performance of the backbone: it introduces only a $0.010$ decrease on VBVR-Bench and a $0.3$-point decrease on RULER-Bench, while reducing the generation cost from $87$ s to $14$ s and from $98$ s to $18$ s, respectively. 
This result suggests that effective video reasoning can be retained in a few-step Reasoner, which provides an efficient backbone for the proposed online optimization.

On VBVR-Bench, the proposed method improves the baseline by $0.115$ overall, from $0.666$ to $0.781$, with consistent gains on both ID ($+0.111$) and OOD ($+0.119$) tasks. In comparison, at comparable test-time cost, Pass@5 provides only a $0.017$ improvement, while ``VLM-as-Solver'' method VideoTPO decreases the overall score by $0.032$. This gap is particularly pronounced on VBVR-Bench, where the structured prompts already specify detailed task rules and target outcomes. Consequently, prompt refinement provides limited additional supervision, whereas the proposed method directly optimizes visual execution under the given rules. We do not evaluate PE on VBVR-Bench because the benchmark already provides carefully designed prompts that explicitly specify the task rules and target outcomes.

On RULER-Bench, the proposed method raises the average score of the baseline Reasoner from $46.4$ to $68.2$, yielding a $21.8$-point improvement. In contrast, PE, VideoTPO, and Pass@5 yield improvements of only $1.9$, $3.9$, and $2.7$ points, respectively. More importantly, the proposed method consistently improves performance across all $30$ evaluated task categories, whereas PE and VideoTPO decrease performance on $7$ and $4$ categories, respectively. Prompt-space methods remain effective on several tasks whose intended outcomes can be clarified through language or commonsense reasoning, such as Festival, Medicine, Life, and hypothetical state changes. However, their benefits are less reliable on tasks that depend on precise visual execution. The proposed method achieves particularly substantial gains on such tasks, including Anomaly, Color, Count, and Direction, indicating that directly optimizing visual reasoning trajectories is more reliable than refining textual specifications alone.

The additional inference-time cost of our method remains manageable due to the efficient adaptation design. On VBVR-Bench, our method costs $69$s per sample, which is comparable to Pass@5 at $70$s and still lower than the original Wan2.2-5B baseline at $87$s. On RULER-Bench, our method costs $88$ s, comparable to Pass@5 at $90$s and lower than the original Wan2.2-5B baseline at $98$s. Under similar or even lower inference cost, the proposed method yields substantially gains than test-time scaling and VLM-as-Teacher paradigm, demonstrating a favorable cost--performance trade-off for per-instance optimization.

\vspace{1mm} \noindent  \textbf{Qualitative Comparisons.}
Figure~\ref{fig:main_vis} presents qualitative comparisons on symbolic and general-purpose video reasoning tasks. Strong closed-source models such as Kling~2.6 can generate visually plausible videos, but it struggles to follow task-specific rules precisely. For example, in the object-moving task (upper-left of Figure~\ref{fig:main_vis}), Kling~2.6 fails to place the objects into the correct dashed targets, and the blue square also changes its shape during the trajectory, violating the process constraint of preserving object identity. 

The step-distilled baseline VGM exhibits more severe reasoning failures due to its smaller model capacity and weaker reasoning priors compared with large-scale closed-source models. In the maze-navigation task (upper-right), the yellow ball splits into multiple instances during the trajectory and therefore violates the entity-consistency constraint. In the anomaly-correction task (lower-right), the generated hand still contains six fingers, showing that the baseline fails to complete the intended correction. 

VideoTPO, which refines the prompt using VLM feedback, does not effectively resolve these issues. In the maze example (upper-right), it still produces invalid intermediate trajectories with duplicated balls, and in the hand-correction example (lower-right), the anomaly remains uncorrected. These examples suggest that prompt refinement alone provides limited help when the main difficulty lies in precise visual execution rather than ambiguous task description.

In contrast, our proposed method consistently satisfies both the final-goal and process-constraint queries synthesized by the VLM Teacher. In the object-moving task (upper-left), it accurately aligns all objects with their corresponding dashed targets while preserving their shapes, colors, and cardinality. In the maze task (upper-right), it guides the yellow ball to the red square without crossing the blocked cells or introducing duplicated instances. In the chair-rotation task (lower-left), it correctly rotates the chair by $90^\circ$ counterclockwise while preserving the chair appearance and keeping the surrounding plant and wall fixed. In the hand-correction task (lower-right), it gradually removes the anomaly and produces a realistic five-finger hand. These examples show that directly optimizing the generated trajectory with process-aware and goal-aware supervision is more effective than relying on fixed generation or prompt-space refinement alone. Additional qualitative examples and video results are provided in our project page.

\subsection{Ablation and Analysis}

We conduct a series of ablation studies to analyze the proposed method from four perspectives. First, we examine the reward design, including the necessity of task-specific online optimization, task-adaptive reward synthesis, and the complementary roles of final-goal and process supervision. Second, we evaluate the efficient adaptation designs that make video-level test-time optimization practical, including step distillation, the choice of denoising step for supervision, the number of sampled frames, the lightweight decoder and the optimization budget by varying the number of online optimization steps. Finally, we analyze the generalization of the proposed method across different VLM Teachers and VGM backbones, and further discuss its remaining limitations. Please refer to our project page for video results.

\begin{table}[!t]
\centering
\caption{Ablation results on VBVR-Bench for fixed online optimization steps, reward design, and efficient adaptation.}
\label{tab:ablation-results}
\small
\setlength{\tabcolsep}{5.0pt}
\resizebox{\columnwidth}{!}{
\begin{tabular}{l | c | cc}
\toprule
\textbf{Variants} & \textbf{Overall} & \textbf{ID Avg.} & \textbf{OOD Avg.} \\
\midrule
\multicolumn{4}{>{\columncolor{color3}}c}{\textbf{Reward Design}} \\
\multicolumn{4}{l}{\textit{w/o Task-specific Online Optimization}} \\
\ \ + Differentiable Reward & 0.688 & 0.716 & 0.660 \\
\ \ + Non-differentiable Reward~\cite{liu2026flow} & 0.681 & 0.707 & 0.655 \\
w/o Task-specific Reward & 0.712 & 0.739 & 0.685 \\
w/o Process Reward & 0.758 & 0.782 & 0.734 \\
w/o Final Reward & 0.692 & 0.718 & 0.666 \\
\midrule
\multicolumn{4}{>{\columncolor{color3}}c}{\textbf{Efficient Adaptation}} \\
w/o Step Distillation & 0.714 & 0.739 & 0.689 \\
w/ Last-step Optimization & 0.705 & 0.713 & 0.698 \\
w/ Full-step Optimization & 0.769 & 0.792 & 0.746 \\
Sample frames = 12 & 0.773 & 0.797 & 0.749 \\
Sample frames = 48 & 0.782 & 0.805 & 0.759 \\
\midrule
Step = 0  & 0.666 & 0.692 & 0.640 \\
Step = 5  & 0.710 & 0.735 & 0.685 \\
Step = 10 & 0.750 & 0.774 & 0.726 \\
Step = 16 & 0.781 & 0.803 & 0.759 \\
Step = 20 & 0.783 & 0.804 & 0.762 \\
Step = 40 & 0.778 & 0.800 & 0.756 \\
\midrule
\rowcolor[HTML]{F4F8EE}
\textbf{Ours} & 0.781 & 0.803 & 0.759 \\
\bottomrule
\end{tabular}
}
\end{table}

\begin{figure*}[!t]
	\centering
	\includegraphics[width=\textwidth]{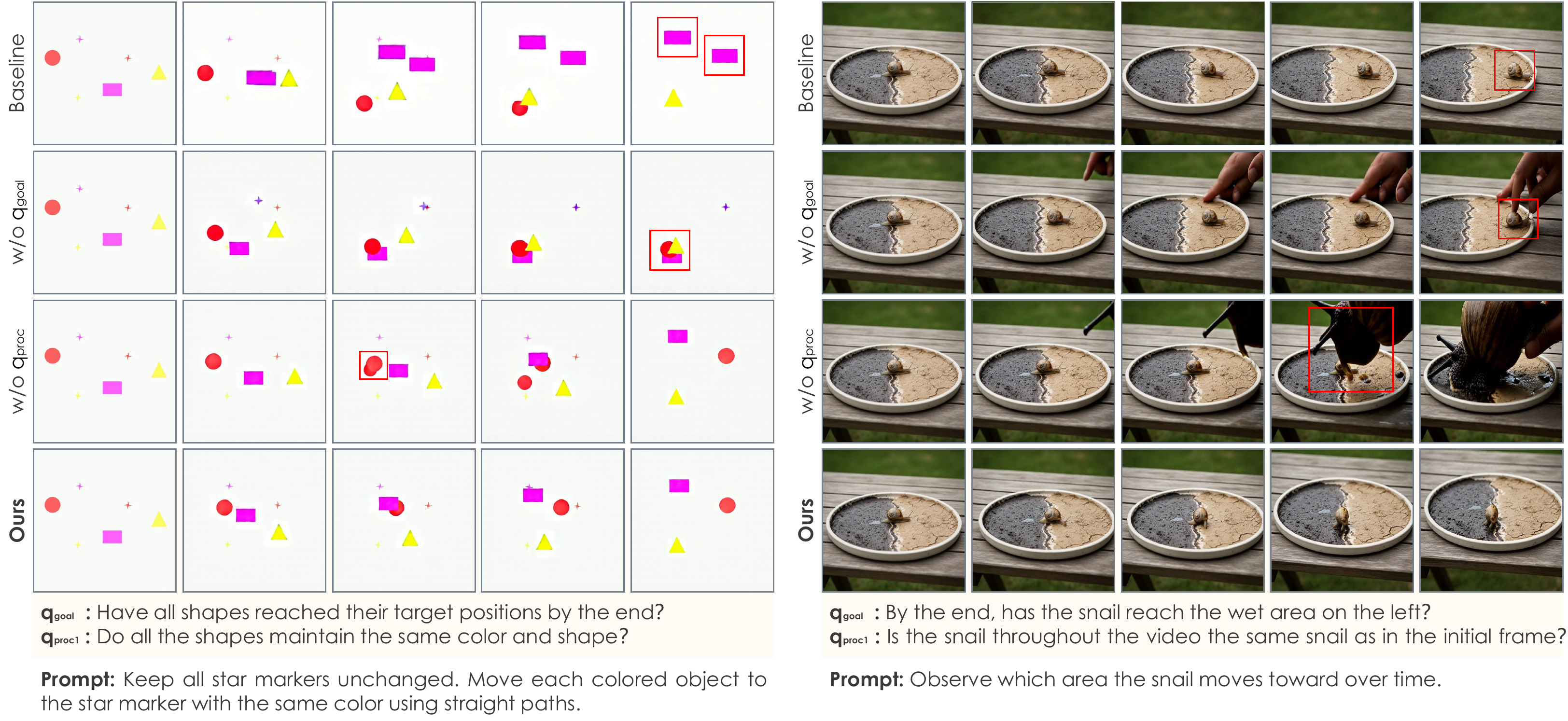}
\caption{Qualitative analysis of reward-query ablations. In the left example, removing process supervision causes the red ball to split despite reaching the target region, while removing final-goal supervision prevents the shapes from reaching their target positions. In the right example, the snail is expected to move toward the moist region on the left; removing process supervision allows a shortcut trajectory by introducing another snail, while removing final-goal supervision fails to guide the snail to the correct region. Key failure areas are marked with red boxes.}
\label{fig:reward_ablation_vis}
\end{figure*}

\vspace{1mm} \noindent  \textbf{Reward Design.}
The first block of Table~\ref{tab:ablation-results} analyzes the design of the proposed supervision mechanism from three aspects. 
First, we examine whether the VLM reward should be used for instance-specific online optimization or shared post-training before inference. 
Replacing the proposed online optimization with shared post-training using differentiable VLM rewards decreases the overall score from $0.781$ to $0.688$. 
Using a non-differentiable reward with Flow-GRPO~\cite{liu2026flow} further decreases the score to $0.681$. 
These results show that simply incorporating VLM feedback during post-training is insufficient; adapting the VGM Reasoner to the rules of each test instance is critical for video reasoning.

Second, we examine the importance of task-specific reward synthesis. 
Replacing the queries synthesized from each task condition with fixed generic queries, which only ask whether the goal is achieved and whether the process is valid, decreases the overall score from $0.781$ to $0.712$. This substantial drop demonstrates that video reasoning requires supervision tailored to each task's specific goals and process constraints, rather than a shared set of generic reward queries.

We set the balance weight between final-goal and process supervision to $\lambda=0.5$ by default, assigning equal importance to task completion and trajectory validity. We ablate the two components of the synthesized supervision by setting $\lambda$ to $1$ and $0$, respectively. Removing process supervision decreases the score from $0.781$ to $0.758$, while removing final-goal supervision leads to a larger drop to $0.692$. These results confirm that the two types of supervision serve complementary roles: final-goal supervision encourages successful task completion, whereas process supervision prevents invalid intermediate trajectories or shortcut solutions.
Fig.~\ref{fig:reward_ablation_vis} provides qualitative evidence for this distinction. 
In the symbolic example, removing final-goal supervision preserves the shapes more consistently during the intermediate process, but fails to guide them toward the required target positions. In the snail-moving example, removing process supervision reaches the target region through an invalid shortcut: a hand introduces another snail rather than moving the original one. 
In contrast, the full reward design satisfies both the intended final goal and reasoning process.

\begin{figure*}[!t]
	\centering
	\includegraphics[width=\textwidth]{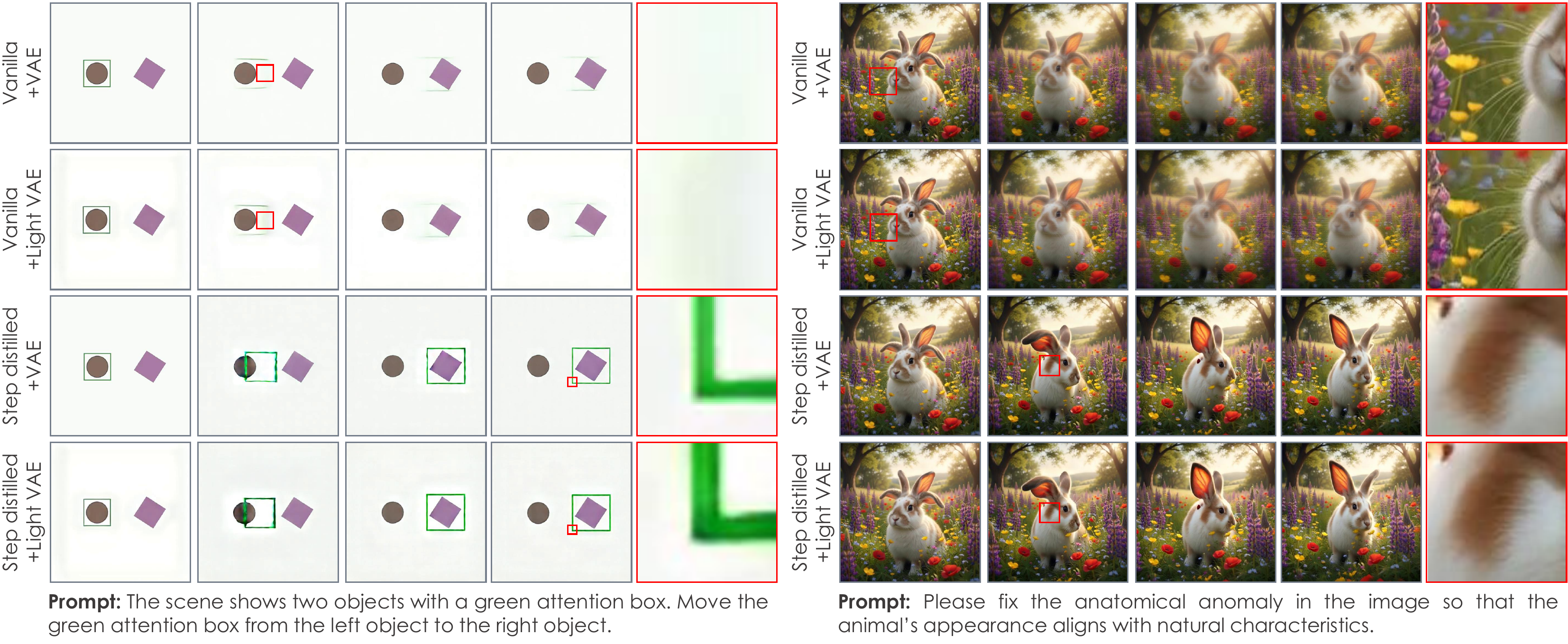}
\caption{Effects of step distillation and lightweight surrogate decoding. Without step distillation, the one-step prediction is blurry and ambiguous in the left symbolic example, and shows almost no visible motion in the right abnormal example, making VLM evaluation unreliable. Step distillation produces more task-informative one-step predictions, while the lightweight surrogate decoder largely preserves the key visual structures needed for supervision. Key differences are enlarged in red boxes.}
\label{fig:efficient_vis}
\end{figure*}

\vspace{1mm} \noindent  \textbf{Efficient Optimization Designs.}
The second and third blocks of Table~\ref{tab:ablation-results} evaluates the key designs that make online optimization efficient and effective. 
Removing step distillation decreases the overall score from $0.781$ to $0.714$. As shown in Fig.~\ref{fig:efficient_vis}, without step distillation, the one-step prediction contains minimal task-relevant motion or state change, making it difficult for the VLM Teacher to judge whether the reasoning task is being completed. In the symbolic example, the decoded square becomes blurry and ambiguous. In the rabbit example, the rabbit remains almost static, providing insufficient visual evidence for evaluating the intended action. In contrast, the step-distilled Reasoner produces a more perceptible one-step approximation of the reasoning trajectory, enabling effective VLM evaluation during online optimization. 

\begin{figure}[!t]
	\centering
	\includegraphics[width=0.40\textwidth]{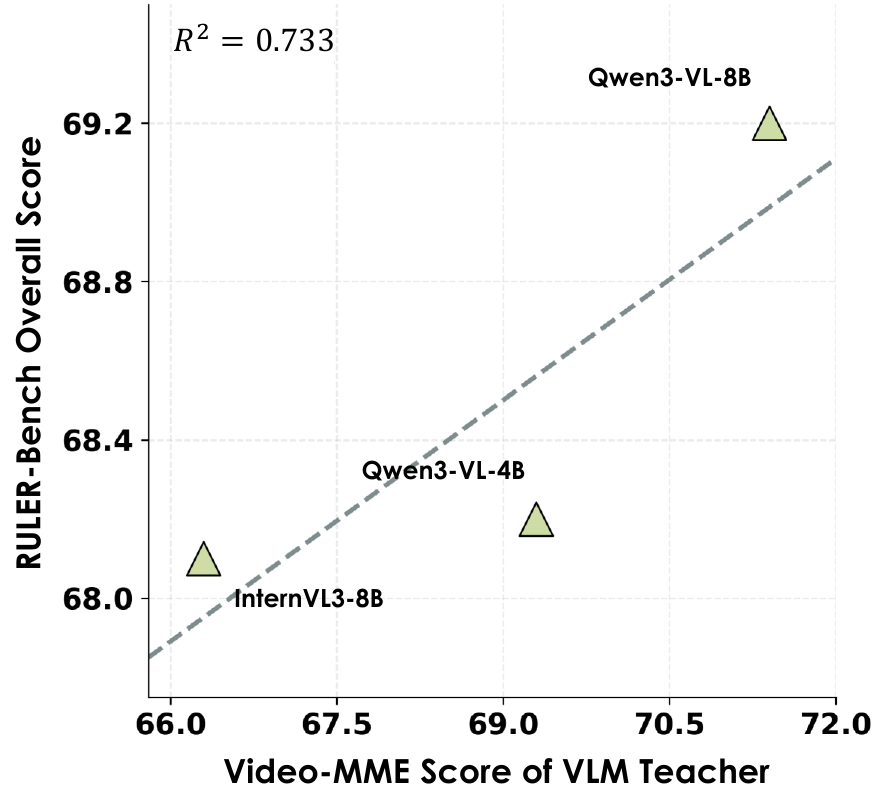}
\caption{Correlation between the video understanding capability of the VLM Teacher, measured on Video-MME, and the resulting performance on RULER-Bench.}
	\label{fig:teacher_correlation}
\end{figure}

We further ablate the denoising step used for optimization. Replacing the proposed first-step optimization with last-step optimization substantially decreases the overall score from $0.781$ to $0.705$. This indicates that optimizing only the final denoising step provides weak supervision for video reasoning, since the high-level motion pattern and task-relevant trajectory are largely determined in early denoising stages, while later steps mainly refine visual details. We also evaluate full-step optimization, where gradients are backpropagated through all four denoising steps. This variant achieves $0.769$, which is still lower than the proposed first-step design. These results suggest that completing and optimizing the full denoising process is not necessary: the early prediction of the step-distilled Reasoner already exposes sufficient reasoning behavior for the VLM Teacher to provide effective supervision, while avoiding the additional cost and potential instability of backpropagating through the full sampling trajectory.

We then study the number of sampled frames used for VLM evaluation. 
Reducing the number of frames from $24$ to $12$ decreases the score to $0.773$, suggesting that overly sparse sampling may miss important intermediate changes. 
Increasing the number to $48$ obtains $0.782$, only $0.001$ higher than the default setting. 
We therefore use $24$ frames as an effective trade-off between reasoning performance and VLM evaluation cost. Fig.~\ref{fig:efficient_vis} further shows that the lightweight surrogate decoder preserves the task-relevant visual structures required for VLM evaluation, despite moderate degradation in visual quality.

Finally, With loss-based early stopping, the proposed method performs only $16$ online optimization steps on average on VBVR-Bench, achieving an overall score of $0.781$ while avoiding unnecessary test-time overhead. To analyze the effective optimization budget, we disable early stopping and evaluate the model with different fixed numbers of online optimization steps. 
As shown in the first block of Table~\ref{tab:ablation-results}, increasing the number of optimization steps from $0$ to $16$ steadily improves the overall score from $0.666$ to $0.781$. 
Extending optimization from $16$ to $20$ steps provides only a marginal gain of $0.002$, while further increasing it to $40$ steps slightly decreases the score to $0.778$. These results indicate that the benefits of online optimization largely saturate after approximately $16$ steps, while excessive optimization may over-optimize the VLM-based objective and introduce visual degradation.

\vspace{1mm} \noindent  \textbf{Generalization across Teachers and Backbones.}
Table~\ref{tab:generalization-results} evaluates whether the proposed framework generalizes across different VLM Teachers and VGM backbones. 
Using Qwen3-VL-4B as the default VLM Teacher, our method achieves an overall score of $68.2$ on RULER-Bench. Replacing it with InternVL3-8B yields a comparable score of $68.1$, while using Qwen3-VL-8B further improves the score to $69.2$. Fig.~\ref{fig:teacher_correlation} shows a strong positive correlation between the video understanding capability of the VLM Teacher, measured by Video-MME performance~\cite{fu2025video}, and the resulting RULER-Bench performance, with $R^2=0.733$. 
This result indicates that the proposed method is compatible with different VLM Teachers, while stronger video understanding generally leads to more effective supervision during online optimization.

\begin{table}[!t]
\centering
\caption{Generalization results on RULER-Bench with different VLM teachers and VGM backbones. $^{*}$ denotes step-distilled model.}
\label{tab:generalization-results}
\small
\setlength{\tabcolsep}{4.0pt}
\resizebox{\columnwidth}{!}{
\begin{tabular}{l | c | ccccc}
\toprule
\textbf{Model Variant} & \textbf{Overall} & \textbf{Humanity} & \textbf{Science} & \textbf{Hypothesis} & \textbf{Semantics} & \textbf{Vision} \\
\midrule
\multicolumn{7}{>{\columncolor{color3}}c}{\textbf{VLM Teacher}} \\
InternVL3-8B~\cite{zhu2025internvl3} & 68.1 & 79.7 & 70.4 & 58.7 & 80.5 & \underline{58.2} \\
Qwen3-VL-8B~\cite{bai2025qwen3} & \textbf{69.2} & \textbf{80.7} & \textbf{71.5} & \textbf{59.5} & \textbf{81.6} & \textbf{59.4} \\
Qwen3-VL-4B~\cite{bai2025qwen3} & \underline{68.2} & \underline{79.9} & \underline{70.6} & \underline{58.9} & \underline{80.8} & 58.1 \\
\midrule
\multicolumn{7}{>{\columncolor{color3}}c}{\textbf{VGM Backbone}} \\
HunyuanVideo-1.5B$^{*}$~\cite{hunyuanvideo2025} & 35.8 & 43.0 & 36.5 & 39.5 & 46.0 & 27.5 \\
\rowcolor[HTML]{F4F8EE}
\ \ \textbf{+ Ours} & \textbf{44.5} & \textbf{51.0} & \textbf{44.0} & \textbf{43.5} & \textbf{54.0} & \textbf{38.3} \\
Wan2.2-5B$^{*}$\cite{wan2025wan} & 46.4 & 55.5 & 47.5 & 50.3 & 58.6 & 36.1 \\
\rowcolor[HTML]{F4F8EE}
\ \ \textbf{+ Ours} & \textbf{68.2} & \textbf{79.9} & \textbf{70.6} & \textbf{58.9} & \textbf{80.8} & \textbf{58.1} \\
\bottomrule
\end{tabular}
}
\end{table}

\begin{figure*}[!t]
	\centering
	\includegraphics[width=\textwidth]{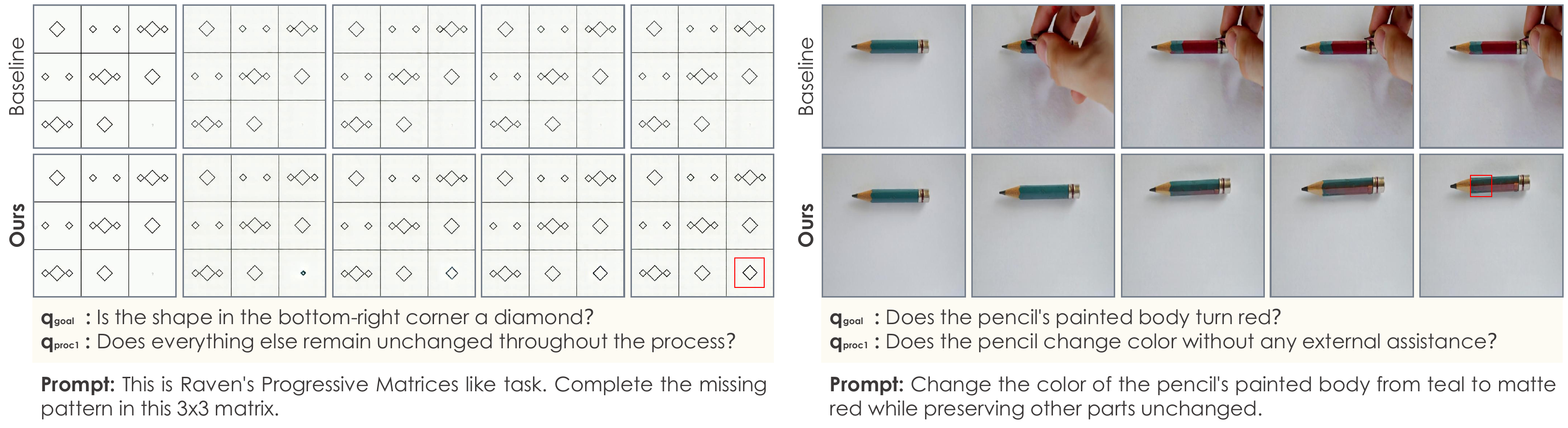}
\caption{Visualization of representative failure and limitation cases. Key failure regions are marked with red boxes.}
	\label{fig:failure_vis}
\end{figure*}

We further evaluate the proposed method with different VGM backbones. 
Applying our method improves the step-distilled HunyuanVideo-1.5B from $35.8$ to $44.5$ on RULER-Bench. The consistent improvements across both backbones demonstrate that the proposed method is not restricted to a specific VGM backbone.

\vspace{1mm} \noindent  \textbf{Failure Cases and Limitations.}
Fig.~\ref{fig:failure_vis} visualizes representative failure cases of the proposed method.
In the RAVEN example, the VLM Teacher synthesizes an incorrect final-goal query by misidentifying the desired final configuration. The correct answer should contain two diamonds, but the synthesized query only checks whether the bottom-right shape is a diamond. As a result, online optimization is guided toward an incomplete objective, even though the generated trajectory satisfies the synthesized query.
In the pencil example, the overall task is largely completed, but the VLM Teacher overlooks a subtle residual error: a small part of the pencil body is not fully transformed into red before the VLM loss falls below the stopping threshold. This illustrates that the proposed supervision can miss fine-grained local errors when they are not sufficiently perceived by the teacher.

To quantify the failure sources, we conduct a human evaluation on 200 generated cases, including 100 cases from VBVR-Bench and 100 cases from RULER-Bench. We count a case as a failure only when it violates the final goal or process constraints of the reasoning task. As shown in Table~\ref{tab:failure_analysis}, the baseline Reasoner fails on 39\% of VBVR-Bench cases and 67\% of RULER-Bench cases. In comparison, the proposed method reduces the failure rates to 18\% and 29\%, respectively, showing that VLM-guided test-time optimization substantially improves task completion and reasoning correctness. We further categorize the failure sources of the proposed method. Most remaining failures are caused by VLM perception errors, accounting for 16\% and 22\% of all evaluated cases on VBVR-Bench and RULER-Bench, respectively. These errors occur when the synthesized queries are correct, but the VLM Teacher overlooks fine-grained visual violations during evaluation. In contrast, incorrect reward-query synthesis accounts for only 2\% and 7\% of the cases, suggesting that the teacher usually derives reasonable task-specific supervision queries. These results indicate that the main limitation of the proposed method lies in the perception granularity of the VLM Teacher, rather than in the online optimization process itself.

\begin{table}[t]
\centering
\caption{Human evaluation of failure sources on 200 cases, including 100 cases from VBVR-Bench and 100 cases from RULER-Bench. A case is counted as a failure only when it violates the final goal or process constraints. Ratios are computed over the 100 evaluated cases in each benchmark.}
\label{tab:failure_analysis}
\small
\setlength{\tabcolsep}{4.5pt}
\resizebox{\columnwidth}{!}{
\begin{tabular}{llcc|cc}
\toprule
\multirow{2}{*}{\textbf{Method}} 
& \multirow{2}{*}{\textbf{Failure Source}} 
& \multicolumn{2}{c|}{\textbf{VBVR-Bench}} 
& \multicolumn{2}{c}{\textbf{RULER-Bench}} \\
\cmidrule(lr){3-4} \cmidrule(lr){5-6}
& & \textbf{Count} & \textbf{Ratio} & \textbf{Count} & \textbf{Ratio} \\
\midrule
\multirow{1}{*}{Baseline} 
& Overall & 39 & 39\% & 67 & 67\% \\
\midrule
\multirow{3}{*}{Ours} 
& VLM perception error & 16 & 16\% & 22 & 22\% \\
& Incorrect reward-query synthesis & 2 & 2\% & 7 & 7\% \\
& Overall & 18 & 18\% & 29 & 29\% \\
\bottomrule
\end{tabular}
}
\end{table}

We also observe a mild visual-quality trade-off. As shown in the qualitative results on VBVR-Bench in Fig.~\ref{fig:main_vis}, the proposed method generally produces visually plausible trajectories, but the optimization process may occasionally introduce slight artifacts or reduce low-level visual fidelity. This is expected because our objective optimizes VLM-level rule satisfaction rather than pixel-level reconstruction quality. Importantly, such visual-quality degradation does not necessarily indicate task failure when the final goal and process constraints are satisfied. As reported in Table~\ref{tab:benchmarking-results}, the proposed method substantially improves the VBVR-Bench overall score from $0.666$ to $0.781$, demonstrating stronger reasoning correctness in terms of final-goal achievement and process-constraint satisfaction. To quantify the visual-quality side effect, we report Fréchet Video Distance (FVD)~\cite{unterthiner2019fvd}. On VBVR-Bench, FVD slightly increases from $21.90$ to $23.37$; since lower FVD is better, this indicates a mild degradation in visual fidelity compared with the baseline Reasoner. Future work may further reduce this trade-off by incorporating lightweight visual-quality regularization during test-time optimization.

\section{Conclusion}
\label{sec:conclusion}

In this work, we introduce a VLM-as-Teacher paradigm for rule-based video reasoning, shifting the role of VLMs from producing textual solutions to supervising visual execution. Specifically, a VLM Teacher synthesizes task-specific reward queries that assess process-constraint satisfaction and final-goal achievement, and provides differentiable feedback to guide a VGM Reasoner through test-time online optimization. Together with efficient adaptation designs, the proposed method enables instance-specific refinement of visual reasoning trajectories at practical test-time cost. Extensive experiments on the symbolic VBVR-Bench and the general-purpose RULER-Bench demonstrate consistent improvements across diverse reasoning tasks, yielding a 16.7-point average performance gain over the baseline Reasoner and substantially outperforming VLM-as-Solver and Best-of-N scaling strategies at comparable test-time cost. These results 
highlight the potential of using VLMs as test-time teachers to bridge high-level logic and visual execution in generative video reasoning.

\section*{Acknowledgement}
This work was supported by Kuaishou Technology and a grant from the NSFC/RGC Collaborative Research Scheme sponsored by the Research Grants Council of the Hong Kong Special Administrative Region, China and National Natural Science Foundation of China (Project No. CRS-HKUST605/25).

{
\small
\bibliographystyle{IEEEtranN}
\bibliography{main}
}

\end{document}